\pdfoutput=1
\documentclass[11pt,a4paper]{article}
\usepackage{times,latexsym}
\usepackage{url}
\usepackage[T1]{fontenc}
\usepackage{arabtex}
\usepackage{amsmath}
\usepackage{utf8}
\usepackage{graphicx}
\setcode{utf8}
 \usepackage[acceptedWithA]{tacl2018v2}
\usepackage[]{tacl2018v2}

\usepackage{xspace,mfirstuc,tabulary}

\newif\iftaclinstructions
\taclinstructionsfalse 
\iftaclinstructions
\renewcommand{\confidential}{}
\renewcommand{\anonsubtext}{(No author info supplied here, for consistency with
TACL-submission anonymization requirements)}
\newcommand{\instr}
\fi

\iftaclpubformat 

\else

\fi

\title{BERT-DRE: BERT with Deep Recursive Encoder for Natural Language Sentence Matching}

\author{Ehsan Tavan$^{1,*}$,$\:$ Ali Rahmati$^{1,*}$,$\:$ Maryam Najafi$^{1,}$\Thanks{Equal contribution. Listing order is random.}$\:$,
        $\:$ Saeed Bibak$^1$,
        $\:$ Zahed Rahmati$^2$ \\
        $^1$ NLP Department, Part AI Research Center, Tehran, Iran \\ 
        \texttt{\{ehsan.tavan, ali.rahmati, maryam.najafi, saeed.bibak\}@partdp.ai}\\
        $^2$ Department of Mathematics and Computer Science, Amirkabir University of Technology, Tehran, Iran\\
        \texttt{zrahmati@aut.ac.ir}
}

\date{}

\begin{document}
\maketitle
\begin{abstract}
This paper presents a deep neural architecture, for Natural Language Sentence Matching (NLSM) by adding a deep recursive encoder to BERT so called BERT with Deep Recursive Encoder (BERT-DRE). Our analysis of model behavior shows that BERT still does not capture the full complexity of text, so a deep recursive encoder is applied on top of BERT. Three Bi-LSTM layers with residual connection are used to design a recursive encoder and an attention module is used on top of this encoder. To obtain the final vector, a pooling layer consisting of average and maximum pooling is used. We experiment our model on four benchmarks, SNLI, FarsTail, MultiNLI, SciTail, and a novel Persian religious questions dataset. This paper focuses on improving the BERT results in the NLSM task. In this regard, comparisons between BERT-DRE and BERT are conducted, and it is shown that in all cases, BERT-DRE outperforms only BERT. The BERT algorithm on the religious dataset achieved an accuracy of 89.70\%, and BERT-DRE architectures improved to 90.29\% using the same dataset.
\end{abstract}

\section{Introduction}
Natural Language Sentence Matching (NLSM) is a fundamental task in Natural Language Processing(NLP) designed to identify similarities in terms of semantic and conceptual content between two input sentence. A wide range NLP tasks, such as Natural Language Inference(NLI), Paraphrase Identification, Question Answering, and Machine Comprehension (MC), are implemented with NLSM. The NLI, also known as Recognizing Textual Entailment (RTE), seeks to determine whether a hypothesis can be deduced from a premise. This requires examining the semantic similarity between the hypothesis and the premise. In Paraphrase Identification, the goal is to determine if two texts are paraphrases or not \cite{lan_neural_2018}.

In Machine Comprehension, the model must match the sentence structure of the passage to the question, pointing out where the answer is located \cite{lan_neural_2018}. Question answering systems rely on NLSM in two ways: (1) question retrieval, (2) the answer selection. In question retrieval, the matching between the query and existing questions, and in answer selection, the matching between the query and existing answers are determined. In fact, answer selection is used to discover the relationship between the query and answer and ranks all possible answers. These NLSM-based question answering systems can be applied in question-answering forums. Many websites use these question-answering forums and they can use these systems to answer their user’s questions. There are so many repetitive questions in these forums that NLSM-based question answering systems can use FAQs to answer users questions. Users will see the answers similar questions, when a new question is asked in the forum. Using this NLSM-based question answering systems user questions will be answered faster \cite{singh_tomar_neural_2017}. Multiple datasets can be accessed in NLSM, including Stanford Natural Language Inference (SNLI) \cite{bowman-etal-2015-large}, Multi-Genre Natural Language Inference (MultiNLI) \cite{williams-etal-2018-broad}, FarsTail \cite{amirkhani2021farstail}, SciTail \cite{AAAI1817368}, and more. One of the challenges when using these datasets is that in many existing samples the two input texts have the same and many similar words. In fact, both texts are very similar in appearance but express different meanings, or conversely, two texts have many different words, but the nature and meaning of both questions are very similar.

The objective of NLSM models is to predict a category or scale value for a pair of input texts, which indicates their similarity or relationship. To achieve this goal, NLSM models generally use two main steps: (1) designing a model to obtain a proper representation of whatever text will be analyzed so that it can extract semantic features from it. (2) By using the representation obtained from texts, a matching mechanism can be developed to extract complex interactions.

In the current NLSM field, deep neural networks are the preferred approach. To determine semantic features and relationships among sentences, researchers use convolutional and recurrent neural networks. The structure of convolutional neural networks (CNNs) \cite{lecun1999object} make them very capable of extracting local features. To this end, CNN have been extensively used in various areas of natural language processing. By considering natural language texts as a sequence of words, it is crucial to extract and analyze temporal features. However Input sequences can be analyzed using recurrent neural networks with their unique structures to extract temporal features. The Long Short Term Memory (LSTM) \cite{hochreiter1997long} a variant of RNN's that is capable of extracting long-term dependency in appropriate way. It has achieved very good results in many NLP tasks by using the LSTM structure like named entity recognition \cite{cho2020combinatorial}, question-answering \cite{othman2019manhattan}, emoji prediction \cite{tavan2020persian} etc. In order to extract text features, deep learning networks require a lot of labeled data, which is one of the challenges of using deep learning, especially in natural language processing. To extract the best textual features today, pre-trained language models are used, which have been trained using a large amount of data. The advantage of these pre-trained language models is that they do not need labeled data to be trained, so they can be trained on a large amount of unlabeled data and then fine-tuned them in downstream tasks. A technique known as transfer learning is one of the best ways to resolve the challenges of natural language processing in low-resource languages today.

In this paper, we used the following steps to design a model based on NLSM principles identifying the semantic similarity of the input text pair:
\begin{enumerate}
\item Collecting Persian Religious question matching dataset for training and evaluating the proposed NLSM model.
\item Investigating related models to the field of NLSM.
\item Implementing ERT with Deep Recursive Encoder (BERT-DRE) model by adding recursive encoder module to BERT \cite{devlin2019BERT}.
\item Evaluating BERT-DRE with related models using the introduced religious and benchmark datasets.
\end{enumerate}
Dataset annotated for this study includes 18,000 samples, contains two questions, appropriate answers, and a match or not-match label for each question pair, which is used for designing a chatbot. This dataset is collected by crawling religious questions from religious question-answering websites, and then by annotating two similar questions are annotated for each question, and dissimilar questions are generated automatically. It was noted our BERT-DRE model, which used the annotated religious dataset to train and evaluate, was able to achieve an F1-score of 90.27\% on the test data, making it the strongest model among the ones studied. Furthermore, in order to better evaluate the BERT-DRE model, the model was trained using SNLI, MultiNLI, FarsTail, and SciTail and achieved appropriate F1-scores.

\section{Related Works}
Natural language processing studies have focused on NLSM, an essential part of many natural language tasks. In this regard, various datasets and models have been developed. There have been different models developed for NLI, semantic matching, and paraphrase identification. Since all of these tasks are similar, the same model can be used today. In most cases, deep neural networks are used for this purpose. To this end, in the following, we shall examine the relevant works.

As stated in \cite{bimpm_ref}, in contrast to previous work performed in text-similarity that adapted the sentence from one direction or used a word-to-word or sentence-to-sentence correspondence only, this study used bilateral multi-perspective matching from several perspectives. As a result, one of the main architectural pillars is based on Bidirectional LSTM. Furthermore, they examined the mechanism of different methods, such as attention. This interaction between aggregate words is then examined from multiple perspectives. In \cite{bimpm_ref} (BiMPM), sentences are matched bidirectionally with a new function calculating similar vectors. This method can result in similarity in multiple ways, which is called multi-perspective cosine similarity.

Language objects internal structures and interactions of language objects need to be properly modeled to achieve a successful matching algorithm. A component of this goal is achieved in \cite{Hu2014ConvolutionalNN} through combining ideas from neural networks' implementation in image and audio processing. Besides demonstrating the hierarchical structures of sentences in layers, this model also maintains rich matching patterns on different levels. 
The \cite{he-etal-2015-multi} (MPCNN) presented work based on CNN, analyzing multiple perspectives of sentences. CNN first extracts features for each sentence, then compares representations of the sentences at various levels of granularity, and then uses various pooling techniques. Following that, multiple similarity metrics are used to compare sentences at multiple levels of granularity. The MPCNN model processes each sentence independently, and there is no interactivity until the final Fully-Connected layer, which results in the loss of a great deal of useful information. For this reason, \cite{Cao2018SemanticMU} and \cite{he-etal-2016-umd} make changes to the MPCNN architecture.

In \cite{Cao2018SemanticMU} has also improved MPCNN, one of which is the use of pre-trained embedding instead of random embedding. In the next step, characters are used to extract features. Finally, an input layer based on attention is added between the embedding and the multi-perspective sentence modeling layers. Attention-based neural network models have been successfully used in answer selection, which is an important sub-task of question answering. These models often represent the question using a vector and match it by referring to the candidate answers. However, questions and answers may be interrelated in complex ways that cannot be represented by single-vector representations. In \cite{10.1145/3209978.3210009}, the idea of using multipurpose attention networks (MANs) is proposed, which aims to discover this complex relationship for ranking question and answer pairs. Unlike previous models, this paper does not turn the question into a single vector, but focuses on different parts of the question from several vectors to show its general meaning, and applies different stages of attention to learning the representations of the candidate answers.

The study \cite{Zhao2019InteractiveAF} proposes an interactive attention model for semantic text matching that uses the global matching matrix to learn representations for source and target texts during interactive attention. This model can take a rich representation of source and target texts and derive an entirely related encoding. In \cite{chen-etal-2017-enhanced}, a model with a good performance called ESIM has a non-complex structure of LSTM layers and is cited in many later works as a base model. It has been claimed that recursive architecture can achieve good performance in local inference modeling and inference composition.

In \cite{tay-etal-2018-compare}, a deep learning model named CAFE is presented in which alignment pairs are compared, compressed, and then spread to the upper layers to increase representation learning. Then, the alignment vectors are expressed in scalar features through factorization layers. One of the other models that was able to achieve good accuracy in SNLI and MultiNLI datasets \cite{yin-etal-2016-abcnn} is another deep learning model that is intended for modeling sentence pairs with CNNs named ABCNN. In this study, three different models of the combination of convolutional layers along with the attention mechanism have been used and the results of these three models have been examined.

Based on BERT representation of sentences and using convolution layers and the semantic role labeler as features added to the model, \cite{Zhang2020SemanticsawareBF} has achieved good accuracy on MultiNLI and QNLI \cite{wang-etal-2018-glue} data. Using a combination of CNN, attention mechanism, and residual connection structures for inference, a deep learning method was proposed in \cite{yang-etal-2019-simple} named RE2. The authors regard attention to pointwise features, previous alignments, and contextual features as three success principles. This model has the advantage of being fast in calculating model inferences. In comparison with other existing models, this model increases speed by six times and uses fewer parameters.
In the \cite{kim2018semantic} a deep learning model is proposed using the LSTM structure named DRCN. In this paper, the alignment module is used to extract the relationship between the words of the first and second text sequences. Residual connections are also used to transfer the extracted features to each layer. The embedding layer in this model includes trainable word embedding, non-trainable word embedding, character embedding, and exact match feature. In this model, due to the use of residual connections, the dimensions of the extracted feature along with the network increase. In order to reduce the feature dimensions in this model, an autoencoder is used.

\section{Data Collection}
In this study, a question matching dataset is annotated to design a religious chatbot. We use this dataset to train and evaluate the BERT-DRE model. In order to collect this dataset, first, 5,000 questions and answers have been crawled from religious question-answering websites. After crawling this data, four annotators aware of religious issues were used to annotate similar questions. Then for each question, two annotators wrote similar questions. The final dataset consists of 10,000 questions with similar questions and related answers.

One of the reasons for finding the dissimilar question was to avoid the random selection of samples. For this reason, the selection of relatively similar samples was done by cosine similarity using TFIDF vectors. According to the evaluations, the best score for this similarity was considered between 10\% and 20\%. Because the results of our experiments showed that values with a similarity score below 10\% are random and values above 20\% have high similarity. As a result, the sample selected with a negative label is not necessarily similar, but it is not completely random. There were about 18,000 pairs of sentences, of which 13,000 were in the train set, 2,700 in the test set, and 2,300 in the validation set. Table \ref{tab:label1} show some samples from collected data. We refer to this collected dataset as a Religious dataset.

\begin{table}
\centering
\begin{tabular}{p{1.3in}|p{1.2in}|p{0.1in}} 
     \hline
     
     \tiny\textbf{First Question}
     &\tiny\textbf{Second Question}
     &\tiny\textbf{Label} \\
     \hline
     
     \tiny\<آيا آفتاب از مطهرات است \\اگر هست شرايط مطهريت آن چيست؟> 
     &\tiny\<آفتاب جزو چيزهای پاک کننده است؟> 
     &\tiny match \\
     \hline
     
     \tiny\<فرق آب کر و آب جاری \\در تطهیر چيست؟>
     &\tiny\<آیا بين غسل و وضو تفاوتی از \\جهت جاری شدن آب بر \\بدن وجود دارد؟>
     &\tiny not match \\
     \hline
     
     \tiny\<آیا طلبکار بايد زکات \\طلب خود را بپردازد؟> 
     &\tiny\<آيا بر طلبکار واجب است که \\زکات طلب خود را بپردازد؟>
     &\tiny match \\ 
     \hline
\end{tabular}
\caption{Samples from collected data}
\label{tab:label1}
\end{table}

To investigate the BERT-DRE, we made experimentation on the well-known datasets as well. We listed the experimental datasets as follows:
\begin{itemize}
    \item \textbf{SNLI} dataset contains 570,000 annotations in three categories: entailment, contradiction, and neutral.
    \item \textbf{MultiNLI} dataset contains 430,000 sample pairs of sentences in ten different genres.
    \item \textbf{FarsTail} dataset is the first entailment dataset was created in Persian language. This dataset includes 10,367 pairs of samples in three categories: entailment, contradiction, and neutral, which are collected from 3539 questions. For each question, three questions (one for each class) were created by the annoters.
    \item \textbf{SciTail} dataset is another entailment dataset created from multiple-choice science exams and web sentences. This dataset contains 27,026 examples with 10,101 examples with entailment label and 16,925 examples with neutral labels.
\end{itemize}

The Table \ref{tab:table2} provides statistical information of these datasets.

\begin{table}[]
    \centering
    \begin{tabular}{cccc}
         \hline \textbf{Dataset} & \textbf{\# of train} & \textbf{\# of dev} & \textbf{\# of test} \\ \hline
          SNLI & 550,152 & 10,000 & 10,000 \\
          MultiNLI & 392,707 & 10,000 & 10,000 \\
          FarsTail & 7,266 & 1,564 & 1,537 \\
          SciTail & 23,596 & 2,126 & 1,304 \\
          Religious & 13,103 & 2,313 & 2,721 \\ \hline
    \end{tabular}
    \caption{Statistical information of datasets}
    \label{tab:table2}
\end{table}

\section{Model}
\begin{figure*}[t!]
    \centering
    \includegraphics[width=6in, height=3in]{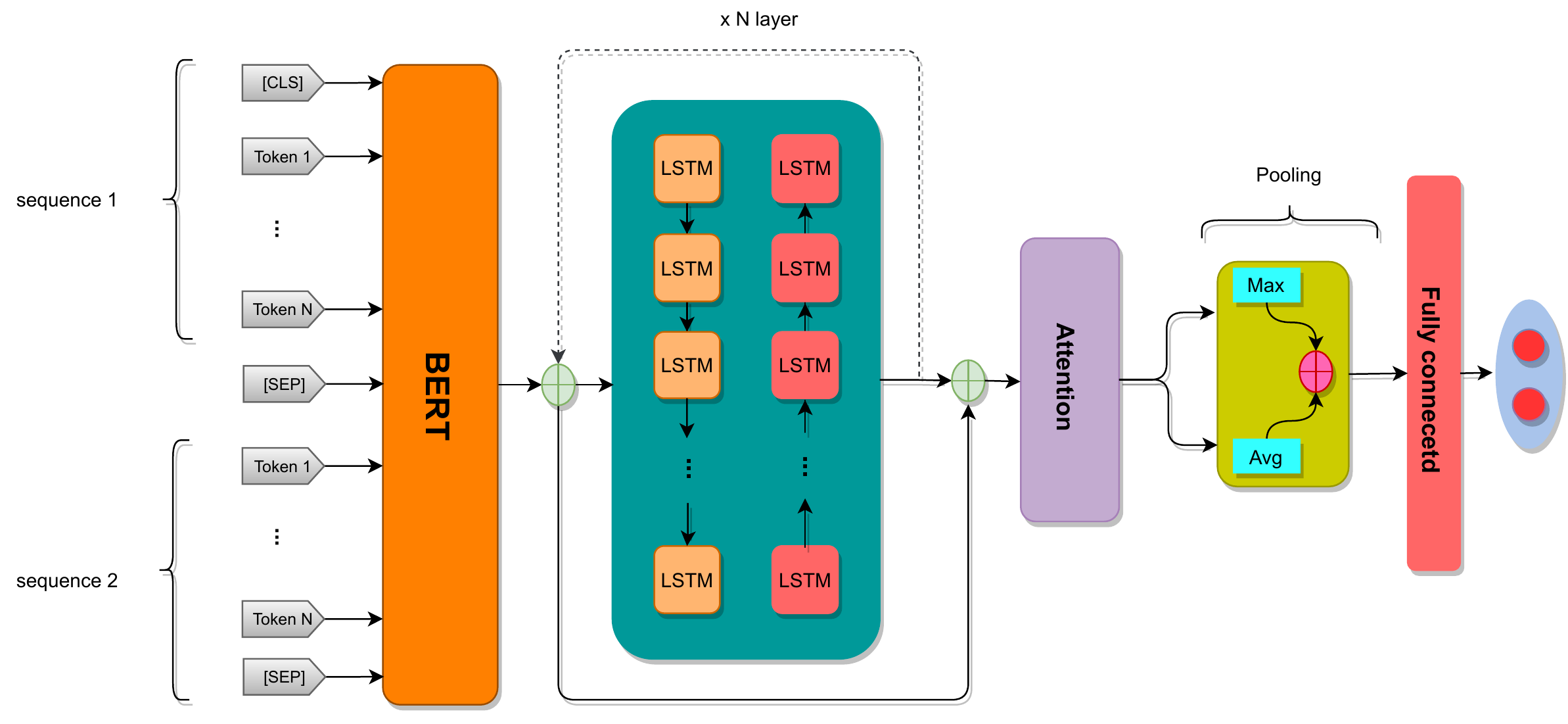}
    \caption{High level overview of BERT-DRE architecture}
    \label{fig:fig1}
\end{figure*}

In this section we provide a layer-by-layer description of BERT-DRE model architecture. Figure \ref{fig:fig1} illustrates a high level overview of our proposed model architecture. BERT-DRE is composed of the following five components:
\begin{enumerate}
    \item Word Representation Layer     
    \item Recursive Encoder Module
    \item Attention Module
    \item Pooling Layer
    \item Prediction Layer
\end{enumerate}

BERT-DRE accepts two input sentences Q (input question) and P (pair question) and estimate the probability distribution $Pr(y \mid Q,P)$ as matching score between Q and P.

\subsection{Word Representation Layer}
This layer aims to learn $k$ dimensional representation for each token. Researchers have used pre-trained language models to obtain these representations. Training these models with huge amounts of data allows them to obtain the embedding vector of each word according to the ones around it. Furthermore, these models can be fine-tuned for any dataset and downstream task, allowing them to achieve good results on various datasets. multilingual BERT and ParsBERT \cite{parsBERT2021} has been used in this study. In this model $Q = \{q_1, …, q_I\}$ and $P = \{p_1, …, p_J\}$ are the model inputs, where $q_i | p_j$ is the $i$th and $j$th token of the $Q | P$ sentence and $I | J$ is the sequence length of $Q | P$. BERT's input is similar to $\{[CLS], q_1, …, q_I, [SEP], p_1, …, p_J, [SEP]\}$. As a result, each of the input tokens becomes a vector with a dimension of 768, which is in fact, a vector representation of each token.

\subsection{Recursive Encoder module}
\begin{figure}[h!]
    \centering
    \includegraphics[width=2.5in, height=3.5in]{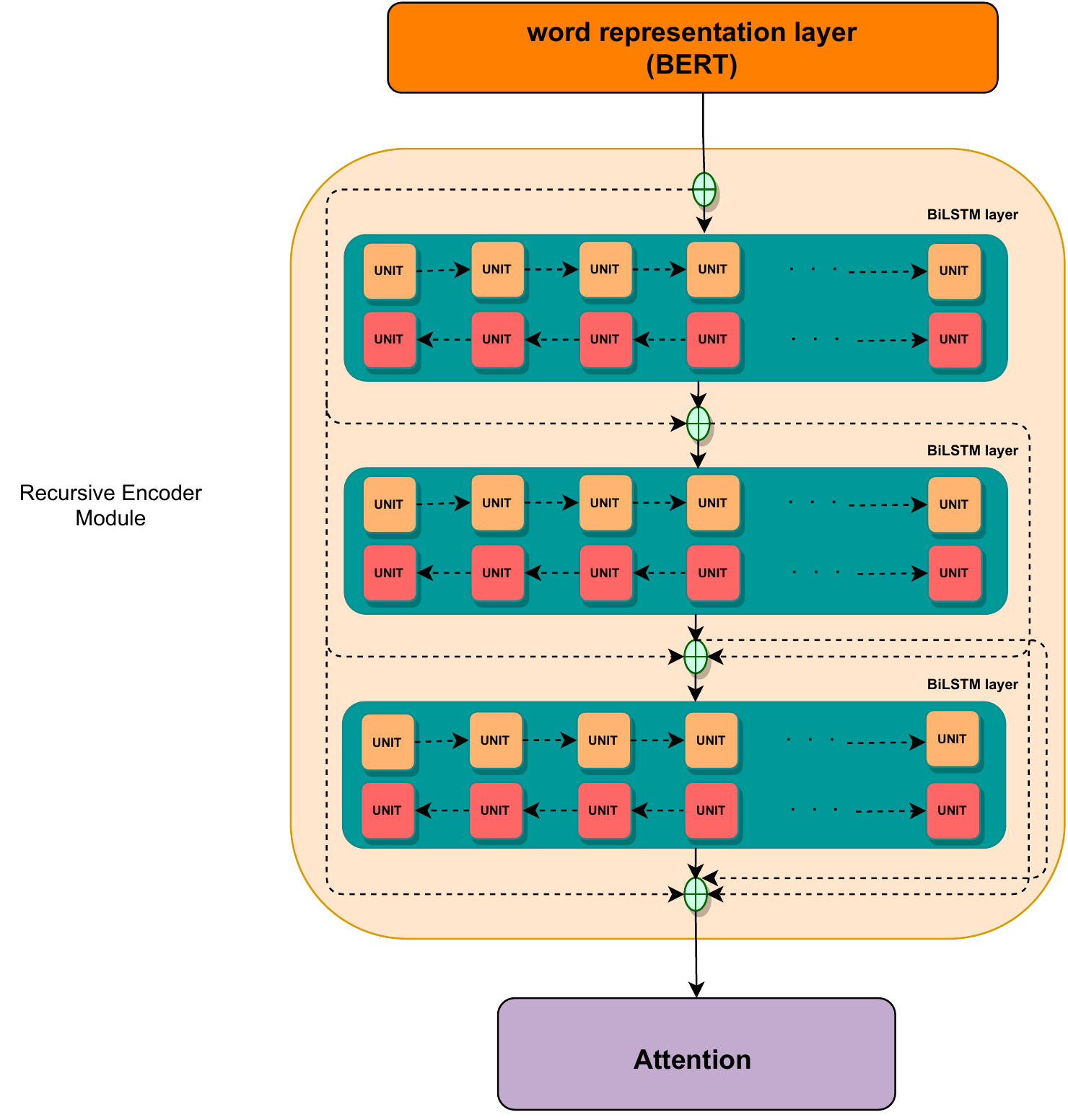}
    \caption{Recursive encoder module structure}
    \label{fig:fig2}
\end{figure}

The recursive encoder module is designed using LSTM layers and residual connection. This module extracts the optimal feature from the output vector of the word representation layer. The LSTM network captures conceptual dependencies very efficiently due to its recurrent nature. The extended structure of recurrent neural networks, known as LSTM, is more efficient in extracting long sequence features than traditional neural networks, and they are developed to eliminate gradient vanishing and exploding in traditional RNN networks. Each sequence can be processed both in the forward and backward by using a bidirectional structure. One of the reasons why three layers of Bi-LSTM were used in the introduced recursive encoder module was to address this issue. the Figure \ref{fig:fig2} illustrates the recursive encoder module structure.

By using residual connections, the information of previous layers can be combined with the information of the present layer and in fact, better and deeper features are disseminated in the Bi-LSTM layers, which can extract the best features in each layer. The idea of disseminating information is that instead of having the previous layer as the input of each  layer, all previous layers are inputs to each  layer. Hence, the model, is able to learn richer features. This concept of information dissemination assumes that the deeper the network, the less information is vanishes because each layer has direct access to the previous layers. The recursive encoder module is describe in Equation \ref{eq:1} and Equation \ref{eq:2}.

$$\overrightarrow{h_i}=\overrightarrow{LSTM}\left(\overrightarrow{h_{i-1}},x_i\right)\ \ \ \ \ \ i=1,\dots ,\ M$$
\begin{equation}\label{eq:1}
        \overleftarrow{h_i}=\overleftarrow{LSTM}\left(\overleftarrow{h_{i+1}},x_i\right)\ \ \ \ \ \ i=M,\dots ,\ 1
\end{equation}

$$H_i=[\overrightarrow{h_i};\overleftarrow{h_i}]$$

$$\textrm{Layer}_1:\ \ x_i={emb}_i$$ 

\begin{equation}\label{eq:2}
    \textrm{Layer}_2:\ x_i=[{emb}_i;\ h^{l-1}_i] 
\end{equation}

$$\textrm{Layer}_3:\ x_i=[{emb}_i;\ h^{l-1}_i,h^{l-2}_i]$$

Where in  Equation \ref{eq:1} $h_i$ is LSTM’s hidden state in $i$th time steps, $x_i$ is LSTM’s input in $i$th time steps and $M$ is sequence length. Equation \ref{eq:2} has shown each LSTM layer's input vector. Where $emb_i$ is embedding vector for $i$th tokens and $l$ is present layer.

\subsection{Attention Module}
The attention module assigns different weights to tokens in order to achieve a better prediction of the desired class, since they all play different roles in the prediction. To this end, the proposed model uses the recursive encoder module, followed by the attention module, which calculates the importance and relationship of the tokens. By calculating the importance of the tokens according to their labels, this module extracts more valuable features from the input sequence. Using the proposed model, the input vector of this module is composed of all three outputs of the Bi-LSTM layers of the recursive encoder module as well as the embedding vectors of the words derived from BERT. The attention module consists of the following components:

$$e_t = \tanh (h_t w_a) $$
\begin{equation}\label{eq:3}
    a_t = \frac{exp(e_t)}{\sum_{i=1}^{T}exp(e_i)}
\end{equation}

For time step $t$, $h_t$ equals token representation, while $w$ equals attention module weights. At each time step, the attention importance score, $a_t$ is computed by multiplying the representations with the weight matrix and normalizing to construct a probability distribution over the words. The representation vector for the text, $v$ is determined by summing attention importance scores across all time steps as weights.

\subsection{Pooling Layer}
We used the pooling module on top of the attention module's output to learn the general representation. The max-pooling and avg-pooling modules are concatenated in pooling layer.
\begin{equation}
    X =[Max([h_1, ... , h_l]) ; Avg([h_1, ... , h_l])]
\end{equation}

where $X$ is the vector representation of the pooling section in $k$ dimensions. Due to the use of this layer, a gradual decrease has been observed in the representation size, which will reduce the number of parameters, as well as model size, as a result, it prevents the model from overfitting. Additionally, they can predict which features of the text are likely to play a significant role in its classification.

\subsection{Prediction Layer}
The aim of this layer is to evaluate the probability distribution $Pr(y\mid Q, P)$. for this purpose, we employ two layer feed-forward neural networks and the softmax function in the output layer. The final prediction layer is computed as follows:

\begin{equation}
    y_{pred} = softmax(w_o * y_o + b_o)
\end{equation}

Where $W_o \in R^{h*n}$, $b_o \in R^n$ and $n$ is the number of classes. The network then trained using cross entropy loss.

\section{Experiments}
We experimented our proposed model on four NLU benchmarks including SNLI, MultiNLI, SciTail, FarsTail and Religious dataset. The proposed model has been compared to many state-of-the-art models, including deep learning and BERT models, and we have compare the results of  BERT and  our BERT-DRE model.

\subsection{Implementation Details}
We implemented our model using PyTorch and trained it on Nvidia V100 GPUs. Training the model is performed by using the Adam optimizer function with an estimated learning rate of 2$e$-5. Following the recurrent layers, dropout is used with a retention rate of 0.5. The batch size is tuned among [32,64,128]. The number of layers of LSTM tuned 1 to 5. Furthermore, each layer has a hidden size between [64, 128, 256]. Depending on the sample length in the dataset, the maximum length will vary for different datasets. For SNLI, MultiNLI, and SciTail datasets, the maximum sentence length is 100. Since FarsTail has long samples, we have set a maximum length of 200 for each sample. Of course, the longest samples in the collected religious dataset are used in this article, in which the maximum length for each sample is 250. 

\subsection{Experimental Results}
The results of the NLSM models and our proposed model on the Religious test set is illustrated in Table \ref{tab:religious}. It has been demonstrated that BERT outperform other NLSM models. Furthermore, the proposed model achieved an accuracy of 90.29\% on test set, which was an improvement of more than 5\% compared to some other deep learning models such as CAFE. As expected, the proposed model has also been able to significantly improve on BERT language model itself, leading to an indication that BERT improves when combined with other models such as recurrent layers. BERT's output can be processed using a deep architecture with promising accuracy compared to other architectures.

\begin{table}[h]
    \centering
    \begin{tabular}{cc}
         \hline \textbf{Model} & \textbf{Accuracy(\%)} \\ \hline
         ESIM \cite{chen-etal-2017-enhanced} & 85.68 \\
         CAFE \cite{tay-etal-2018-compare} & 83.67 \\
         RE2 \cite{yang-etal-2019-simple} & 86.28 \\
         DRCN \cite{kim2018semantic} & 84.88 \\
         BiMPM \cite{bimpm_ref} & 72.80 \\
         ABCNN \cite{yin-etal-2016-abcnn} & 73.04 \\
         BERT \cite{devlin2019BERT} & 89.70 \\
         BERT-DRE & \textbf{90.29} \\
        \hline
    \end{tabular}
    \caption{Experimental results on Religious test set}
    \label{tab:religious}
\end{table}

The results of the SNLI dataset are shown in Table \ref{tab:snli}. The results of BERT-DRE on the test set were compared to those from other state-of-the-art models. The BERT-DRE model, with an accuracy of 90.83\%, achieved a competitive and close accuracy with other models in this dataset. We adjusted BERT model on this dataset to have a fair comparison. As a result of this fine-tuning, the accuracy of BERT model was 91.00\%, which is close to models that used BERT in their structure, such as SemBERT \cite{Zhang2020SemanticsawareBF} with 91.90\% and SJRC \cite{zhang2019explicit} with 91.30\% accuracy. By combining BERT and deep learning model, BERT-DRE model was also able to achieve close accuracy to the mentioned hybrid models. On the other hand, it performs better than many models such as RE2 and CAFE and improves by 2.33\% and 1.93\%, respectively.

\begin{table}[h]
    \centering
    \begin{tabular}{cc}
         \hline \textbf{Model} & \textbf{Accuracy(\%)} \\ \hline
         ESIM \cite{chen-etal-2017-enhanced} & 88.00 \\
         CAFE \cite{tay-etal-2018-compare} & 88.5 \\
         BiMPM \cite{bimpm_ref} & 87.5 \\
         DRCN \cite{kim2018semantic} & 88.9 \\
         RE2 \cite{yang-etal-2019-simple} & 88.9 \\
         SemBERT \cite{Zhang2020SemanticsawareBF} & \textbf{91.9} \\
         MT-DNN \cite{liu-etal-2019-multi} & 91.6 \\
         SJRC \cite{zhang2019explicit} & 91.3 \\
         Ntumpha \cite{liu-etal-2019-multi} & 90.5 \\
         BERT \cite{devlin2019BERT} & 91.00 \\
         BERT-DRE & 90.83 \\ 
        \hline
    \end{tabular}
    \caption{Experimental resutls on SNLI test set}
    \label{tab:snli}
\end{table}

The results of the FarsTail test set are shown in Table \ref{tab:farstail}. In \cite{amirkhani2021farstail}, it is shown that the use of language models such as BERT has been able to achieve an accuracy of 83.38\% in test set. The BERT-DRE model with an improvement of 0.63\% has achieved an accuracy of 84.01\%. Considering the improvement, it can be shown that BERT-DRE has been able to perform relatively better than BERT by using a deep recursive encoder module.
\begin{table}[h]
    \centering
    \begin{tabular}{cc}
         \hline \textbf{Model} & \textbf{Accuracy(\%)} \\ \hline
         DecompAtt\cite{parikh-etal-2016-decomposable} & 66.62 \\
         HBMP\cite{Talman2019SentenceEI} & 66.04 \\
         ESIM \cite{chen-etal-2017-enhanced} & 71.16 \\
         BERT \cite{devlin2019BERT} & 83.38 \\
         BERT-DRE & \textbf{84.01} \\ 
        \hline
    \end{tabular}
    \caption{Experimental resutls on FrasTail test set}
    \label{tab:farstail}
\end{table}

MultiNLI and SciTail results are also shown in Tabels of \ref{tab:MultiNLI} and \ref{tab:sciail}. With the MultiNLI dataset, the proposed model has been able to achieve state-of-the-art results with a favorable difference. As shown in this comparison, the BERT-DRE model is 1.44\% better than BERT in the mismatch data set. In addition, the proposed model was able to achieve a significant improvement over both CAFE and ESIM models. The proposed model achieved an accuracy of 91.50\% on the SciTail dataset, which is close to state-of-the-art models.
\begin{table}[h]
    \centering
    \begin{tabular}{ccc}
         \hline \textbf{Model} & \textbf{Match}  & \textbf{Notmatch}\\ \hline
         ESIM \cite{chen-etal-2017-enhanced} & 72.4 & 72.1 \\
         CAFE \cite{tay-etal-2018-compare} & 80.2 & 79.0 \\
         MT-DNN-ensemble\\ \cite{Liu2019ImprovingMD} & \textbf{87.9} & \textbf{87.4} \\
         MT-DNN\\ \cite{liu-etal-2019-multi} & 86.7 & 86.0 \\
         BERT \cite{devlin2019BERT} & 86.7 & 85.9 \\
         BERT-DRE & 86.07 & 87.34 \\
        \hline
    \end{tabular}
    \caption{Performance comparison on MultiNLI (accuracy)}
    \label{tab:MultiNLI}
\end{table}

\begin{table}[h]
    \centering
    \begin{tabular}{cc}
         \hline \textbf{Model} & \textbf{Accuracy(\%)} \\ \hline
         CA-MTL\cite{pilault2021conditionally} & 96.84 \\
         MT-DNN \cite{liu-etal-2019-multi} & \textbf{94.14} \\
         RE2 \cite{yang-etal-2019-simple} & 86.04 \\
         CAFE \cite{tay-etal-2018-compare} & 83.34 \\
         ESIM \cite{chen-etal-2017-enhanced} & 70.64 \\
         BERT \cite{devlin2019BERT} & 92.04 \\
         BERT-DRE & 91.54 \\
        \hline
    \end{tabular}
    \caption{Experimental resutls on SciTail test set}
    \label{tab:sciail}
\end{table}

Table \ref{tab:fscore_datasets} summarizes the obtained F1-scores by the BERT-DRE model on the train, validation and test sets separately for each benchmark.

\begin{table*}[h]
    \centering
    \begin{tabular}{cccccc}
         \hline \textbf{} & \textbf{Religious} & \textbf{SNLI} & \textbf{MultiNLI} & \textbf{FarsTail} & \textbf{SciTail} \\ \hline
        Train & 94.69 & 91.45 & 82.41 & 84.01 & 96.44 \\
        Test & 90.27 & 90.82 & 87.34 & 83.86 & 91.60 \\
        Dev & 84.26 & 91.31 & 86.07 & 82.61 &93.47 \\
        \hline
    \end{tabular}
    \caption{Performance comparison (F1-score) on datasets}
    \label{tab:fscore_datasets}
\end{table*}

\begin{table*}[h]
    \centering
        \begin{tabular}{cccccc} \hline 
        \textbf{Id} & \textbf{Parameters} & \textbf{FarsTail} & \textbf{Religious} & \textbf{MultiNLI(Match)} & \textbf{MultiNLI(Mismatch)} \\ \hline 
        1a & 1Layer-Lstm & 83.38 & 88.50 & 81.09 & 82.00 \\  
        1b & 2Layer-Lstm & 81.43 & 89.38 & 81.48 & 81.90 \\  
        1c & 3Layer-Lstm & \textbf{83.86} & \textbf{90.27} & \textbf{87.34} & \textbf{86.07} \\ 
        1d & 4Layer-Lstm & 83.82 & 90.00 & 86.95 & 85.67 \\
        1e & 5Layer-Lstm & 83.10 & 89.21 & 85.38 & 85.39 \\ \hline 
        2a & 3Lstm-64Hidden & 81.82 & 88.76 & 85.93 & 85.53 \\ 
        2b & 3Lstm-128Hidden & \textbf{83.86} & \textbf{90.27} & \textbf{87.34} & \textbf{86.07} \\ 
        2c & 3Lstm-256Hidden & 82.94 & 88.89 & 85.91 & 85.57 \\ \hline 
        3 & 3Lstm\_no\_residual & 81.39 & 89.67 & 85.73 & 85.24 \\ \hline 
        \end{tabular}
    \caption{Ablation study results on dev sets on the corresponding datasets (F1-score has been calculated)}
    \label{tab:ablation_table}
\end{table*}

\subsection{Ablation Study}
\begin{figure}[h]
    \centering
    \includegraphics[width=3in, height=2in]{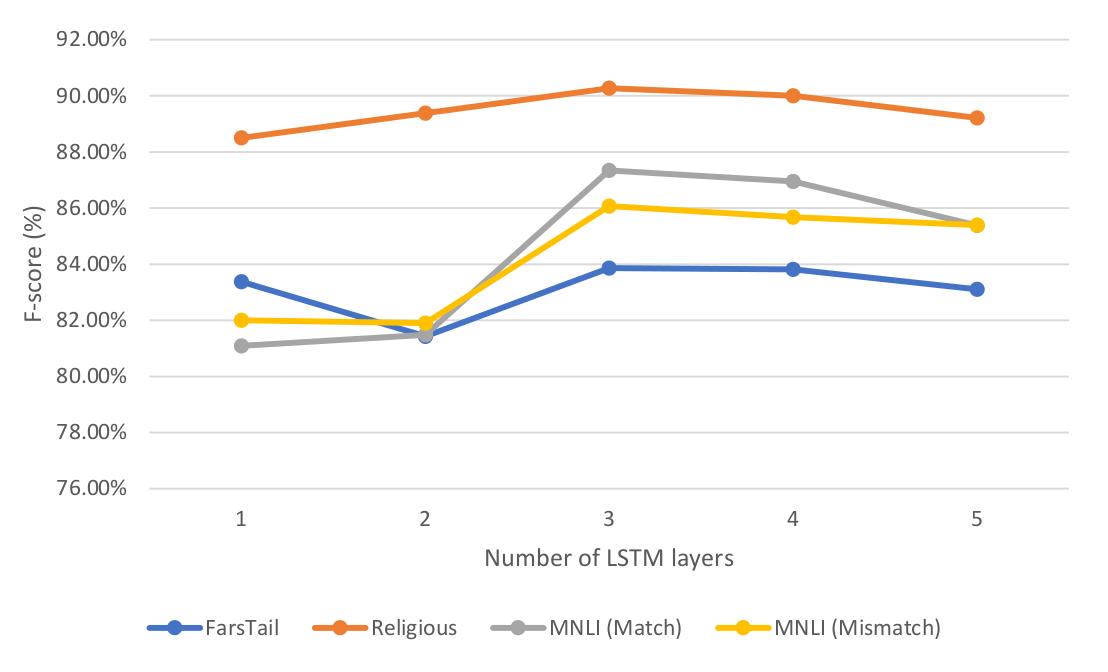}
    \caption{Effect of LSTM layers in an ablation study}
    \label{fig:ablation_lstm_layers}
\end{figure}

We present an ablation study of our model comparing it with three different ablation baselines: (1) investigate the effect of decreasing and increasing the number of LSTM layers within the recursive encoder module, (2) investigate the effect of decreasing and increasing the number of neurons within the recursive encoder module, (3) investigate the effect of using a residual connection on the recursive encoder module. There are three datasets in the ablation study report: FarsTail, Religious, and MultiNLI. The first five lines of the Table \ref{tab:ablation_table}, 1(a) to 1(e), change the number of Bi-LSTM layers from one to five sequential layers to assess the model's performance with different layers. Figure \ref{fig:ablation_lstm_layers} shows the effect of LSTM layers in an ablation study.
As shown in the Table \ref{tab:ablation_table}, three sequential layers of Bi-LSTM had the best performance in all the datasets. There is also some evidence that a large number of layers (3 layers) can improve performance to some extent, but anything beyond that will have no effect on performance. 

\begin{figure}[h]
    \centering
    \includegraphics[width=3in, height=2in]{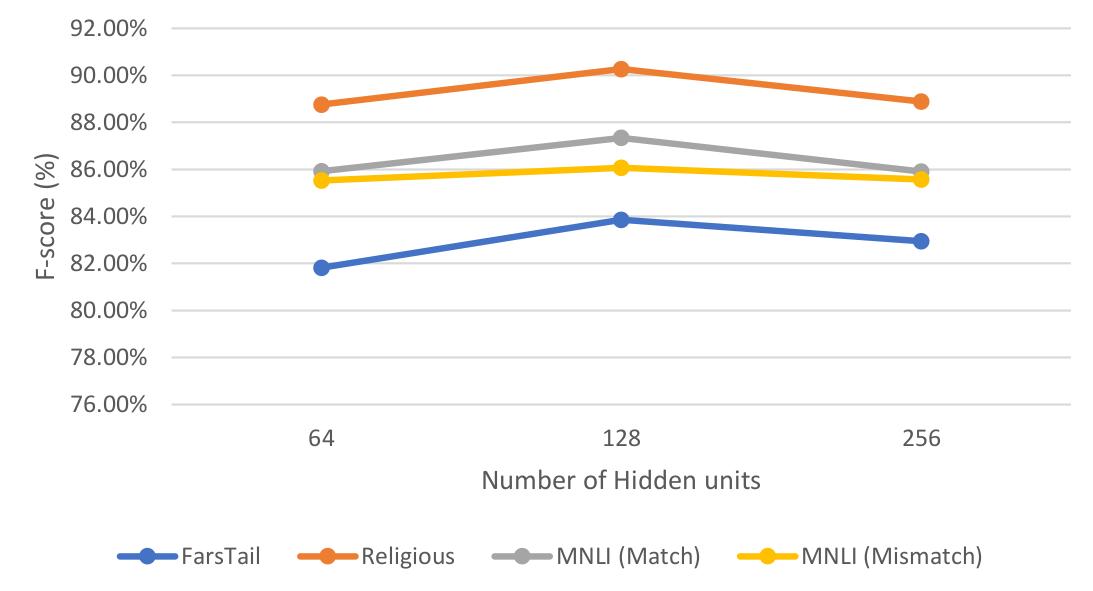}
    \caption{Effect of LSTM units in an ablation study}
    \label{fig:ablation_lstm_units}
\end{figure}

The number of neurons in each LSTM layer was changed to examine model performance in the next step. A choice of 64, 128, or 256 neurons was considered. Finally, all datasets showed that 128 neurons per layer performed better. We found that increasing the number of neurons to a certain number may improve performance in this study. Figure \ref{fig:ablation_lstm_units} shows the effect of LSTM units in an ablation study.

The last step was to investigate the effect of removing the residual connection from the model. In the absence of residual connections, the BERT-DRE model using residual connection was unable to improve its accuracy. 

\section{Conclusion}
In this paper, we proposed a novel architecture by adding recursive encoder module with BERT. The proposed model uses BERT as an embedding layer and On top of the embedding layer, three bidirectional LSTM were used densely to represent the input text semantically. In order to pay more or less emphasis on different words, an attention module is applied to the outputs of LSTMs. This makes the semantic representations more informative. These semantic representations are passed to a pooling layer consisting average and max pooling To make the resulting feature maps more robust to features’ positional changes. To show the potential of this architecture, we evaluated it in four benchmark datasets and got competitive results. The evaluation results in our annotated data show that the accuracy of proposed model is very acceptable. Our empirical results suggest that our proposed model improves the performance on some NLP benchmarks (e.g., FarsTail) with the state-of-the-art pre-trained models (e.g., BERT). This has been developed for the Persian/English language but it could be easily extended to other languages.

\section*{Acknowledgement}
We would like to express our gratitude to the Part AI Research Center for supporting and funding this research and the Technology Engineering team for setting up GPU machines and Hardware infrastructures.

\bibliography{tacl2018v2-template}
\bibliographystyle{acl_natbib}

\end{document}